\pdfoutput=1

\documentclass[11pt]{article}

\usepackage{acl}

\usepackage{times}
\usepackage{latexsym}
\usepackage{amsmath}
\usepackage{amsfonts}

\usepackage[T1]{fontenc}

\usepackage[utf8]{inputenc}

\usepackage{microtype}

\usepackage{inconsolata}

\usepackage{graphicx}

\title{Graphically Speaking: Unmasking Abuse in Social Media with Conversation Insights}

\author{Célia Nouri\textsuperscript{1,2} \quad 
  Jean-Philippe Cointet\textsuperscript{2} \quad 
  Chloé Clavel\textsuperscript{1,3} \\
  \textsuperscript{1}INRIA, ALMAnaCH \\
  \textsuperscript{2}Sciences Po, médialab \\
  \textsuperscript{3}Télécom Paris \\
  \texttt{\{celia.nouri, chloe.clavel\}@inria.fr}, 
  \texttt{jeanphilippe.cointet@sciencespo.fr}
}

\begin{document}
\maketitle

\begin{abstract}
Detecting abusive language in social media conversations poses significant challenges, as identifying abusiveness often depends on the conversational context, characterized by the content and topology of preceding comments. Traditional Abusive Language Detection (ALD) models often overlook this context, which can lead to unreliable performance metrics. Recent Natural Language Processing (NLP) methods that integrate conversational context often depend on limited and simplified representations, and report inconsistent results. In this paper, we propose a novel approach that utilize graph neural networks (GNNs) to model social media conversations as graphs, where nodes represent comments, and edges capture reply structures. We systematically investigate various graph representations and context windows to identify the optimal configuration for ALD. Our GNN model outperform both context-agnostic baselines and linear context-aware methods, achieving significant improvements in F1 scores. These findings demonstrate the critical role of structured conversational context and establish GNNs as a robust framework for advancing context-aware abusive language detection. The code is available at \href{https://anonymous.4open.science/r/ConversationALD-F476/}{this link}.
\end{abstract}

\textcolor{red}{\textbf{Disclaimer:} This paper contains discriminatory content that may be disturbing to some readers.}

\section{Introduction}
\label{sec:introduction}

The expansion of social media has facilitated global communication but has also amplified the spread of abusive language (AL), posing significant challenges \citep{duggan2017onlineharassment, saveski2021structure}. Abusive language refers to communication that demeans, offends, or marginalizes individuals or groups, encompassing hate speech, toxicity, offensive language, and cyberbullying \citep{vidgen2021introducing, bourgeade2024humans}. However, most ALD models classify comments in isolation, disregarding conversational context, which is crucial for accurate classification \citep{pavlopoulos2020toxicity, menini2021abuse, vidgen2021introducing}. 

\begin{figure}[t]
  \includegraphics[width=\columnwidth]{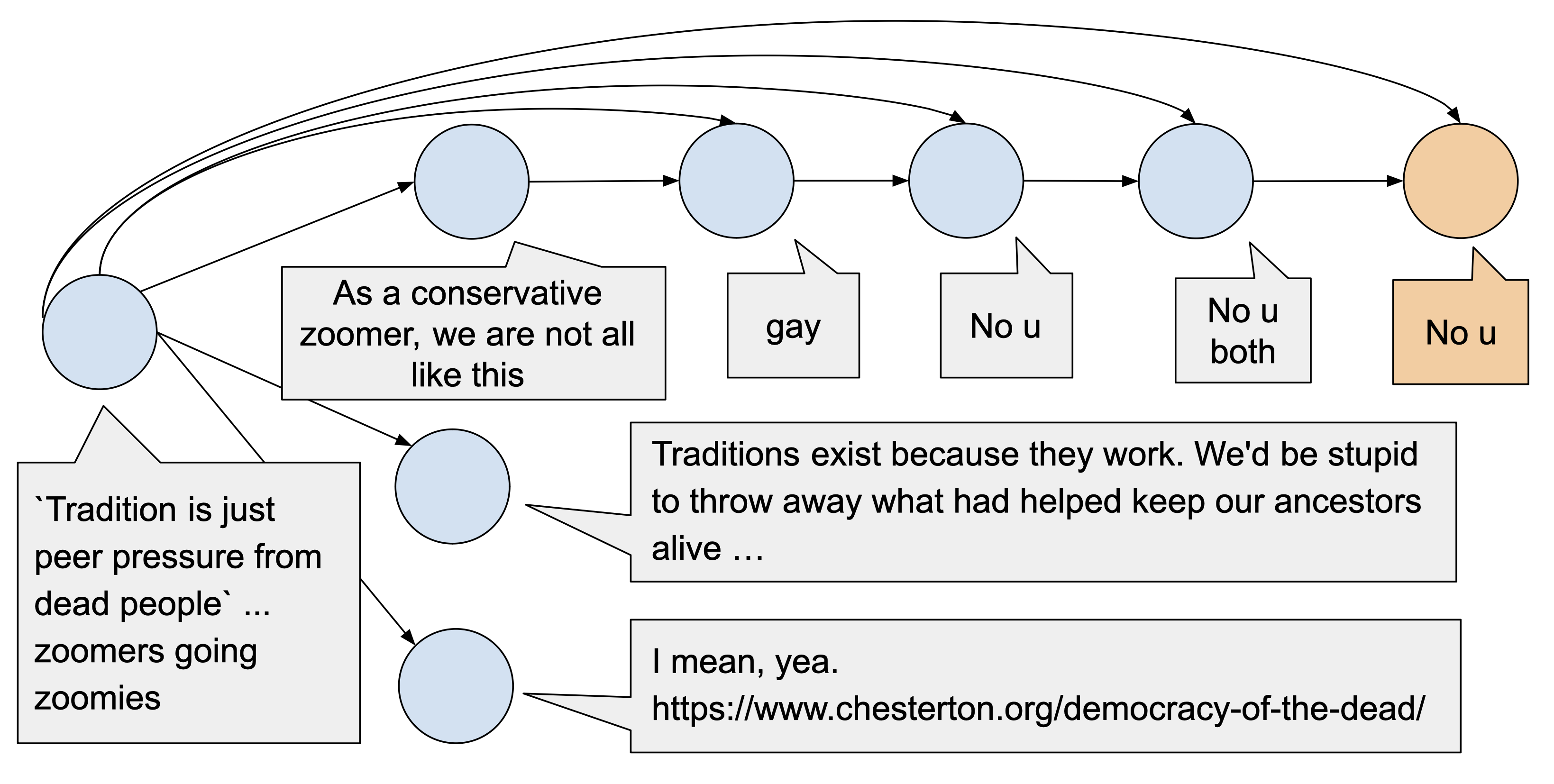}
  \caption{Example conversation from the Contextual Abuse Dataset (CAD), the graph was generated from our Affordance-based method. The target node is labeled abusive and colored in orange.}
  \label{fig:example}
\end{figure}

Figure~\ref{fig:example} illustrates the challenge of context-aware ALD, where the goal is to classify a target comment (in orange) using preceding context without knowing the abusiveness of prior comments. The comment \textit{“No u”} is labeled abusive, but its meaning is unclear in isolation. Examining the full conversation reveals that it occurs within a reply chain initiated by a homophobic insult, triggering a sequence of reactive comments. This pattern exemplifies the snowball effect of abusive speech, where insults propagate and reinforce toxicity. Understanding such interactions requires capturing conversation thread structures and contents beyond solely looking at the immediate preceding comment. 

Despite efforts to incorporate context into ALD, existing methods remain limited. Many define context narrowly, considering only the previous comment or the original post \citep{yu2022hate, ive2021revisiting}. Others treat context as a flat sequence, failing to model the conversation structure, leading to incomplete contextual understanding \citep{bourgeade2024humans}.

To address these limitations, we introduce a graph-based framework for ALD that models conversations as graphs, where nodes represent comments, and edges capture reply relationships. This structure preserves the conversation topology and allows contextual information to propagate across multiple interaction levels. Although our study focuses on Reddit, our approach is relevant to any platform featuring threaded discussions. Through extensive experiments and analyses, we demonstrate that our graph-based models significantly outperform both context-agnostic approaches and prior context-aware models that rely on flat context representations. These results confirm the advantages of explicitly modeling the conversation topology and leveraging graph neural networks to capture contextual dependencies in ALD. 
\paragraph{Contributions} 
Our contributions are as follows:  \\
(1) We propose a graph-based framework for ALD that effectively models Reddit conversations while preserving their structure and relevant content. \\ 
(2) We analyze the optimal amount of conversational context needed for graph-based models to maximize performance. \\ 
(3) We compare our graph-based approach with existing context-aware NLP models, highlighting the strengths and limitations of graph-based ALD.  
\paragraph{Paper Organization}

Section~\ref{sec:related-work} reviews prior work on context-aware ALD. Section~\ref{sec:methodology} presents our methodology, including problem formulation, graph construction, and model architecture. Section~\ref{sec:experiments-results} describes our experimental setup and results, evaluating the impact of different context modeling strategies. Finally, Section~\ref{sec:conclusion} summarizes our findings and outlines directions for future research.

\section{Related Work}
\label{sec:related-work}

This section examines the role of conversational context in human annotations, reviews context-aware ALD datasets, and discusses previous efforts to integrate conversational context into NLP models and graph-based approaches for ALD.  

\subsection{Impact of Context on Human Annotations}

Previous studies have shown that the inclusion of context can significantly alter how comments are perceived and annotated for toxicity or abusiveness.
For instance, \citet{pavlopoulos2020toxicity} investigated the impact of context by annotating 250 Wikipedia Talk page comments under two conditions: in isolation and with context, where context included the post title and the previous comment. They found that 5\% of the labels changed when context was provided, with most changes occurring from nontoxic to toxic. Similarly, \citet{menini2021abuse} re-annotated 8,000 tweets from the Founta dataset \citep{founta2018large} with and without context, where context comprised all preceding messages in the thread. Conversely, their results showed a decrease in the percentage of abusive labels from 18\% to 10\% when context was provided, indicating that annotators perceived fewer tweets as abusive when they had additional contextual information.
These contrasting findings highlight the complexity of incorporating context into ALD and underscore its influence on human perception. Further studies \citep{yu2022hate, vidgen2021introducing} similarly found that providing conversational context impacts the interpretation of abusiveness. These studies emphasized the need for context-aware datasets to improve ALD systems.

\subsection{Datasets}
\label{subsec:datasets}

Numerous datasets have been developed to support abusive language detection and related tasks such as identifying toxicity, hate speech, racism, and sexism \citep{waseem2016hateful, davidson2017automated, golbeck2017large, founta2018large}. However, most of these datasets focus on isolated comment instances, ignoring conversational context during both annotation and modeling. 

While context-aware datasets have been developed across various platforms—including Twitter \citep{menini2021abuse, ihtiyar2023dataset}, Wikipedia, and Reddit \citep{yu2022hate}—they often exhibit limitations such as narrow definitions of context (e.g., considering only the preceding comment or the initial post) \citep{qian2019benchmark, yu2022hate}, small sample sizes \citep{menini2021abuse, pavlopoulos2020toxicity}, or inconsistent annotation quality \citep{hebert2024multi}. We focus our study on Reddit since the platform provides rich and structured conversational threads. Given our research focus, we prioritize datasets that encompass complete Reddit conversation threads, as they provide the necessary depth and structure for analyzing ALD models. Notably, the Contextual Abuse Dataset (CAD) \citep{vidgen2021introducing} aligns with our criteria, offering extensive conversational context essential for our study, and high-quality annotations in context of the entire thread. More details about the CAD dataset will be provided in Section~\ref{sec:dataset}.

\subsection{Context-aware flat models for ALD}

The value of conversational context in ALD has inspired various neural architectures designed to incorporate context into classification tasks. 
A common baseline approach, \textit{Text-Concat}, concatenates the context (e.g., preceding comments or the main post) with the target text, processing the combined input through a transformer like BERT \citep{devlin2019bert}. Studies such as \citet{bourgeade2023what, menini2021abuse, ive2021revisiting} have demonstrated the utility of this method for ALD. Another explored approach is \textit{Embed-Concat}, which embeds the context and target text separately using distinct transformer encoders before combining the embeddings for classification \citep{bourgeade2024humans}. These models serve as baselines in our work, enabling us to compare the performance of our graph-based method. Methods like history embedding \citep{ive2021revisiting} attempt to preserve separate representations of context and target text but face limitations in modeling reply-relationships or multi-turn conversational structures. Moreover, these approaches show inconsistent performance across datasets, underlining the need for more robust techniques to integrate context effectively.  

Large generative language models (GLMs) have shown strong performance in ALD and context-aware ALD through prompting strategies \citep{guo2024investigation, chiu2021detecting}, including Chain-of-Thought (CoT) prompting \citep{wei2022chain} and Few-Shot prompting \citep{brown2020language}. However, these models face critical limitations in real-world scenarios. Proprietary models like GPT-4 (1.76T parameters) lack transparency, making them unsuitable for content moderation, while open-source alternatives such as LLaMA-2-13B (13B parameters) \citep{touvron2023llama}, DeepSeek-V2 (236B total, 21B active) \citep{deepseekv2} or more recently DeepSeek-V3 (671B total, 37B active) \citep{deepseekv3}. Compared to these models, our graph-based approach detailed in Section~\ref{subsec:model-archi}, offers a more parsimonious solution, combining BERT (110M parameters) with graph aggregation (approximately 6M parameters per GAT layer). Our approach offers a fast (100–200 ms—orders of magnitude faster than GLM alternatives that take several sec), computationally frugal and interpretable alternative while preserving conversational structure. 

\subsection{Context-aware Graph Models for ALD}

In this section, we review prior works leveraging graphs to represent online conversations, highlighting differences in graph construction and embedding generation, which are key aspects that set our method apart.

\paragraph{Graph Construction}  

Graph-based approaches for ALD vary in how they represent relationships between social media comments. Some methods construct fully connected graphs, linking messages based on cosine similarity between text embeddings \citep{wang2020sosnet, duong2022hatenet}. While effective for propagating labels across datasets, these approaches overlook conversational structure and fail to prioritize messages within a thread. Other works reconstruct retweet paths using temporal and follower relationships \citep{beatty2020graph}, but these methods are platform-specific and do not generalize well beyond Twitter. Temporal graphs have also been used in chat-based platforms, where context is defined by surrounding messages in the chat \citep{cecillon2021graph, papegnies2019conversational}. However, this approach is unfit for structured threads, such as those on Reddit. More closely related to our work, several studies construct conversation graphs based on reply relationships, with nodes representing comments and edges denoting replies \citep{hebert2024multi, agarwal2023graph, meng2023predicting, zayats2018conversation}. For example, \citet{hebert2024multi} use such graphs but incorporate multimodal embeddings that combine post-image and text features as node attributes. While \citet{zayats2018conversation} also build reply-based graphs for Reddit threads, their goal is to predict the popularity rather than the abusiveness of a comment.
Unlike previous methods, our approach trims conversation graphs to mimic what users see when writing a comment, leveraging the default Reddit rendering settings. 

\paragraph{Context Embedding Generation}

Several methods generating embedding representation from conversation graphs focus on global conversation embeddings that summarize the structural properties of a conversation. For example, \citet{meng2023predicting} apply average pooling to node features across a conversation tree, encoding attributes such as the number of replies and overall tree shape. Similarly, \citet{hebert2024multi, hebert2022predicting} use Graphormer \citep{ying2021dotransformers} to generate embeddings that capture global structural features such as node centrality and connectivity. While these approaches are effective at representing the overall conversation structure, they overlook localized interactions and the specific contextual nuances that can be critical for abusive language detection.
Our work adopts a different perspective by focusing on the local conversation context that users directly interact with, rather than relying on global conversation summaries.
Closer to our work, \citep{agarwal2023graph} propose GraphNLI, which generates context embeddings through random graph walks. Their method uses fixed probabilities to favor paths toward the root and applies discount factors to penalize nodes further from the target comment. In contrast, our approach with GATs dynamically learns the importance of contextual nodes, offering a more flexible and targeted mechanism to capture conversational nuances. 

\section{Methodology}
\label{sec:methodology}

\begin{figure*}[t]
  \includegraphics[width=\linewidth]{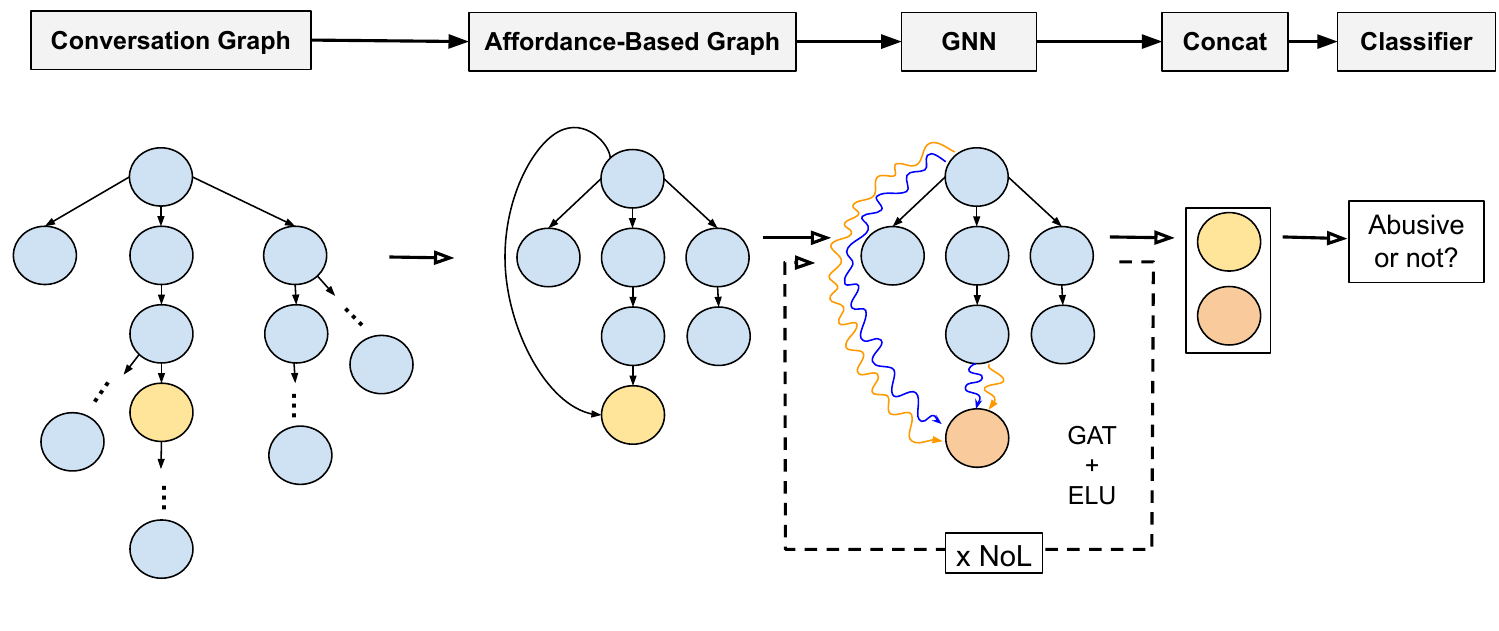}
\caption{Overall Model Architecture. Nodes represent text embedding representations. The yellow node is for our target comment, while the orange node is for the conversation context. \textit{x NoL} stands for \textit{times Number of Layers}. For readability, we did not represent all the edges going from the post node to all other nodes.}
  \label{fig:experiments}
\end{figure*}

This section details the problem formulation, and how it was instantiated for the task of detecting abusive language in Reddit conversations, leveraging graph-based modeling and GNNs.

\subsection{Problem Formulation}

The task of detecting abusive language in social media conversations can be formulated as a binary classification problem. Given a conversation thread \( T \) consisting of \( N \) comments, our objective is to classify whether a specific comment, \( c_i \), within the thread is abusive (\( y_i = 1 \)) or non-abusive (\( y_i = 0 \)), incorporating its conversational context. 

Let \( T \) represent a conversation thread of \( N \) comments: 
\[
T = \{c_1, c_2, \dots, c_N\},
\]  
where \( c_i \) is the \( i \)-th comment to have been posted in the thread \( T \) ordered by posted time. Each comment \( c_i \) has an associated text \( u_i \). 

The thread \( T \) has a graph structure, with comments connected based on reply relationships. This structure is represented as a directed graph:  
\[
G(T) = (V, E),
\]  
where \( V \) is the set of nodes representing comments, and \( E \) is the set of edges representing reply relationships. An edge \( (c_j, c_i) \in E \) exists if \( c_i \) is a reply to \( c_j \). For each node \(v_i \in V\), the feature vector \(\mathbf{x}_i \in \mathbb{R}^d\) is derived from the [CLS] token embedding of its text \(u_i\) using a pre-trained language model, such as BERT. 

For a target comment \( c_i \in V \), the task is to predict its label \( y_i \in \{0, 1\} \), where \( 1 \) denotes \textit{abusive}. This prediction is made using the graph \( G(T) \) and features derived from the comment texts \( u_i \) and the graph structure.

\subsection{Affordance-based Graph Representation}
\label{subsec:context-graph}

Reddit conversations can contain hundreds of comments, but users typically only read top-level replies or preceding comments along the path to their comment of interest. Due to memory constraints and for efficiency reasons, we had to implement a trimming strategy to keep conversation graphs of reasonable size while keeping all relevant context information. We developed a graph trimming affordance-based strategy that aligns with the default Reddit rendering algorithm, which determines the comments visible to a user when writing a reply while adhering to memory constraints. We tried other trimming strategies that did not yield better results, and are presented in Appendix~\ref{fig:trim-strat}.

Specifically, for each target comment \( c_i \), we define the relevant conversational context as subgraph \( G_i \) from \( G(T) \). The subgraph \( G_i = (V_i, E_i) \) includes nodes \( V_i \) and edges \( E_i \) corresponding to comments providing relevant context for \( c_i \). This subgraph is trimmed to align with the default Reddit rendering algorithm, which determines the comments visible to a user when writing a reply. User scores correspond to the number of upvotes minus the number of downvotes for a given comment or post. The subgraph \( G_i \) includes: (i) The original post (\( c_1 \)), (in blue in Figure~\ref{fig:trim-strat}); (ii) The top 5 replies to the root post, ranked by user scores (in green in Figure~\ref{fig:trim-strat}); (iii) The highest-scoring reply to each of these top-5 depth-1 comments (in yellow in Figure~\ref{fig:trim-strat}); (iv) The full reply path leading to the target comment \( c_i \) (in red in Figure~\ref{fig:trim-strat}).

To model user interaction flow, each node connects to the original post. Formally, for a post \( p \in V_i \), we add an edge \( (p, c_m) \) to \( E \) for every \( c_m \in V_i \setminus \{p\} \). We also experimented with a trimmed variant where only the target node \( c_i \) connects to \( p \) via \( (p, c_i) \), but this approach did not improve performance (see Appendix~\ref{subsec:trim-strat}). Figure~\ref{fig:trim-strat} illustrates a conversation trimmed using the affordance-based graph construction method.

\subsection{Model Architecture}
\label{subsec:model-archi}

We employed a Graph Attention Network (GAT) \citep{velickovic2018graph} to model contextual relationships within the conversation graph. For each node \( v_m \) in the graph \( G \), let \(\mathbf{x}_m^{(l)} \in \mathbb{R}^{d_l}\) represent the node embeddings at layer \( l \), where \( d_l \) is the embedding dimension. Each GAT layer updates the node embeddings as follows:  
\begin{equation}
\mathbf{x}_m^{(l+1)} = \text{ELU} \left( \sum_{n \in \mathcal{N}(m)} \alpha_{mn} \mathbf{W}^{(l)} \mathbf{x}_n^{(l)} \right),
\label{eq:gat_layer_update}
\end{equation}
where \( \mathcal{N}(m) \) denotes the neighbors of node \( m \), \( \mathbf{W}^{(l)} \) is a learnable weight matrix, \(\mathbf{a}\) is  a learnable weight vector that calculates the attention scores between nodes by applying it to the combined transformed embeddings of the nodes, and \( \alpha_{mn} \) are attention coefficients computed as:  

\begin{equation}
\resizebox{\columnwidth}{!}{$
\alpha_{mn} = 
\frac{\exp\left(\text{LeakyReLU} \left( \mathbf{a}^T \left[ \mathbf{W}^{(l)} \mathbf{x}_m^{(l)} \, | \, \mathbf{W}^{(l)} \mathbf{x}_n^{(l)} \right] \right) \right)}
{\sum\limits_{k \in \mathcal{N}(m)} \exp\left(\text{LeakyReLU} \left( \mathbf{a}^T \left[ \mathbf{W}^{(l)} \mathbf{x}_m^{(l)} \, | \, \mathbf{W}^{(l)} \mathbf{x}_k^{(l)} \right] \right) \right)}
$}
\end{equation}

After \( L \) GAT layers, the embedding of the target node \( \mathbf{x}_i^{(L)} \) is concatenated with its text embedding \( \mathbf{x}_i \), producing a final representation \( \mathbf{z_i} \in \mathbb{R}^{2d} \). This representation is first passed through a fully connected layer, with parameters \(\mathbf{W}_f\) and \(b_f\), that reduces its dimensionality back to \(d\), the original embedding size of the text model (768):
\begin{align}
\mathbf{h} &= \mathbf{W}_f \mathbf{z} + b_f.
\end{align}
The transformed representation \(\mathbf{h}\) is then passed to the classifier layer, with parameters \(\mathbf{W}_c\) and \(b_c\), of the text model to predict \(y\):
\begin{align}
\hat{y} &= \sigma(\mathbf{W}_c \mathbf{h} + b_c),
\end{align}
where \(\sigma\) denotes the sigmoid activation function.

The model parameters are optimized by minimizing the binary cross-entropy loss. The formulation for the loss is detailed in Appendix~\ref{subsec:objective-function}. Details about hyper-parameters and training setup can be found in Appendix~\ref{subsec:experiment-setup}.

\section{Dataset}
\label{sec:dataset}

\begin{figure}[t]
  \includegraphics[width=\columnwidth]{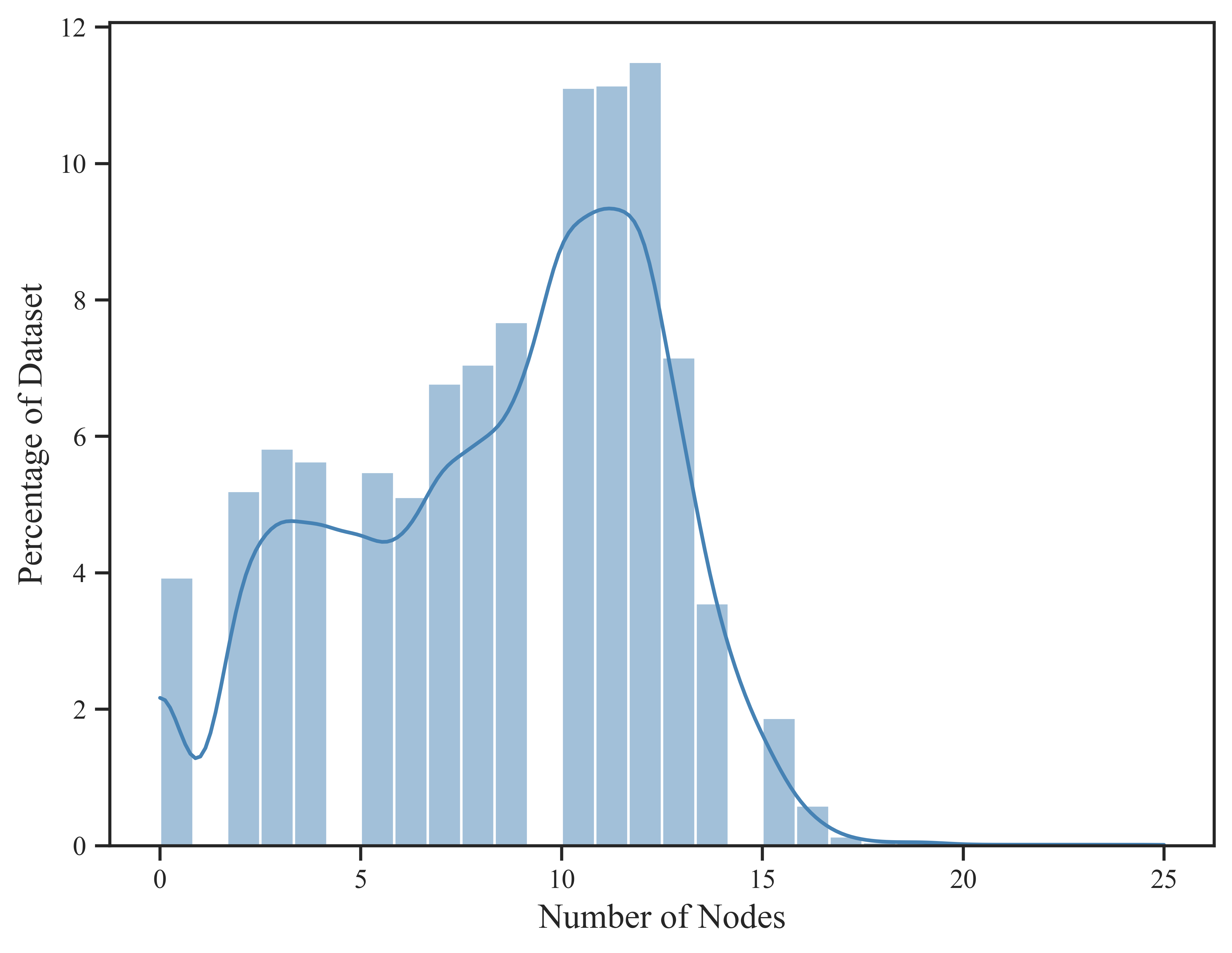}
  \caption{Node Distribution per Graph After Affordance-Based Trimming.}
  \label{fig:graph-stats}
\end{figure}

We use the Contextual Abuse Dataset (CAD) \citep{vidgen2021introducing}, the only high-quality dataset that provides full conversation threads to annotators for abusive speech classification.  
\paragraph{General Description} CAD consists of approximately 25,000 Reddit comments annotated using a target-based taxonomy: Identity-directed, Affiliation-directed, and Person-directed Abuse for the abusive class, and Neutral, Counter Speech, and Non-Hateful Slurs for the non-abusive class. The dataset spans 16 subreddits known for abusive content, with no single subreddit contributing more than 20\% of the data.  \paragraph{Reddit Conversation Description} Reddit discussions are highly structured, with multiple parallel threads. Comment lengths range from brief remarks to over 10,000 words, though 99.3\% fit within the 512-token limit of BERT-based encoders. Conversations also vary in size, with some exceeding 400 comments, while the average training conversation contains 22 comments. Given computational constraints and the fact that users typically see only a subset of a conversation, we applied a trimming strategy based on affordances as described in Section~\ref{subsec:context-graph}. Graphs in our dataset contain maximum \(25\) nodes, and \(9\) at the median. The distribution of the number of nodes per graph is displayed in Figure~\ref{fig:graph-stats}. 
\paragraph{Annotation Process}
CAD employs a rigorous annotation process, combining extensive conversational context, consensus-based adjudication, and expert supervision. Each entry was initially annotated by two trained annotators, with disagreements resolved through consensus adjudication under expert supervision, refining the annotation guidelines when necessary. A final expert review ensured consistency, yielding a Fleiss' Kappa of \(0.583\), indicating moderate agreement. This score compares favorably with other abusive language datasets, considering the complexity of the six-class taxonomy and the inherent subjectivity of ALD \citep{davani2022dealing, vidgen2019challenges, sap2019risk, rottger2022contrasting}.  Annotators had access to the full preceding thread, unlike our model, which only uses textual content and omits visual elements—a direction left for future work. Annotators recorded whether context was critical for classification, which was the case in one-third of abusive instances. We leverage this information to compare our model and baselines on context-sensitive \textit{versus} context-free cases in Section~\ref{subsec:resultsrq2}.  

\section{Experiments and Results}
\label{sec:experiments-results}

To evaluate our approach, we design experiments to address the following research questions: \\
- \textbf{RQ1}: What is the optimal amount of conversational context for GNN models in ALD? \\
- \textbf{RQ2}: How do graph-based approaches compare to other context-aware architectures for detecting abusive language?

This section details our experiments and findings which allow us to answer the two above questions. The experimental setup for all experiments is described in Appendix~\ref{subsec:experiment-setup}.

\subsection{RQ1. Optimal Conversational Context}
\label{subsec:resultsrq1}

\paragraph{Experiments} To determine the optimal conversational context for ALD, we evaluate graph models described in Section~\ref{subsec:model-archi} with 1 to 5 GAT layers, corresponding to 1-hop to 5-hop neighborhoods. Each layer aggregates node features from immediate neighbors, expanding the contextual radius. Formally, the node embedding update equation for each layer is described in Equation~\ref{eq:gat_layer_update}.

\paragraph{Results}

Table~\ref{tab:gat_layers} presents the F1-scores for graph models with different numbers of GAT layers, along with the median and maximum number of nodes in their corresponding receptive fields. These values are derived from the conversation graphs in our experimental dataset. The best performance (\(\text{F1} = 0.7624\)) is obtained with three layers, where the receptive field contains a median of 5 nodes and a maximum of 12 nodes. This finding supports our hypothesis that the limited conversational context typically used in the literature is insufficient for ALD in online discussions. However, increasing the number of layers beyond three does not yield further gains and, in some cases, slightly reduces performance. This stagnation is likely due to the inclusion of less relevant distant comments in wider receptive fields. Moreover, deeper models introduce additional complexity without sufficient training data, potentially leading to overfitting and diminishing returns.  

Notably, the three-hop neighborhood captures most nodes in the affordance-based graphs (see Figure~\ref{fig:graph-stats}), which, by design, are the most contextually relevant to the target comment. While extending context beyond immediate replies improves classification, the performance differences between models with two to five layers are not statistically significant. This suggests that additional conversational context beyond three hops does not provide significant gains in our dataset. Further evaluation on larger and more diverse datasets is necessary to determine whether deeper context windows can enhance performance or if they primarily introduce irrelevant information.

\begin{table}[h!]
    \centering
    \small
    \begin{tabular}{|c|c|c|}
        \hline
        \textbf{GAT Layers} & \textbf{(Max, Median) Nodes} &\textbf{Mean F1 $\pm$ CI} \\
        \hline
        1 & (3, 2) & 0.7537 $\pm$ 0.0069 \\
        2 & (7, 3) & 0.7613 $\pm$ 0.0041 \\
        \textbf{3} & (12, 5) & \textbf{0.7624 $\pm$ 0.0058} \\
        4 & (13, 7) & 0.7592 $\pm$ 0.0065 \\
        5 & (14, 8) & 0.7609 $\pm$ 0.0043 \\
        \hline
    \end{tabular}
    \caption{Mean F1-score (\(\pm\) 95\% CI) for GAT models with different number of GAT layers, averaged over 10 runs. The (Max, Median) Nodes column features the maximum, and median number of comments in the 1-hop to 5-hop neighborhoods.}
    \label{tab:gat_layers}
\end{table}

\subsection{RQ2. Graph-based vs. Flattened Models}
\label{subsec:resultsrq2}

\paragraph{Experiments}

We compare our graph-based models with three baselines. \textbf{No Context} which classifies the target comment using BERT embeddings of the target text without considering the conversational context. 
\textbf{Text-Concat} which concatenates the target comment with preceding comments (trimmed to match the graph model’s context) as a single input sequence, separated by [SEP] tokens. We use Longformer for its 4096-token limit which allows to consider and extended context. Finally, \textbf{Embed-Concat} generates BERT embeddings for each comment, combines context embeddings pairwise through a fully connected layer to form a 768-dimensional vector, adds the target node embedding, and passes the result through a classification layer.

\paragraph{Results}

Table~\ref{tab:baselines} reports the F1-scores for each model. The GAT model (3 layers) achieves the highest performance (\(\text{F1} = 0.7624\)), surpassing all text-based baselines. In particular, the Flattened-Context models (Text-Concat and Embed-Concat) underperform compared to BERT (No Context), reinforcing previous findings \citep{menini2021abuse, bourgeade2024humans} that a naively concatenating context may introduce noise, limiting its utility for ALD.

\begin{table}[h!]
    \centering
    \small
    \begin{tabular}{|c|c|}
        \hline
        \textbf{Model} & \textbf{Mean F1 $\pm$ CI} \\
        \hline
        No Context & 0.7453 $\pm$ 0.0076 \\
        Text-Concat & 0.7417 $\pm$ 0.0081 \\
        Embed-Concat  & 0.7488 $\pm$ 0.0025 \\
        GAT 3L (\textit{ours}) & 0.7624 $\pm$ 0.0058 \\
        \hline
    \end{tabular}
    \caption{Mean F1-score (\(\pm\) 95\% CI) for flattened text-based baselines and graph-based models, averaged over 10 runs.}
    \label{tab:baselines}
\end{table}

To assess model performance in Context-Sensitive Samples (CSS), we examine instances for which annotators explicitly indicated that prior conversational context was crucial for labeling. These context boolean labels are present almost uniquely for abusive samples, and the context-sensitive cases account for approximately one-third of the dataset’s positive cases.

\begin{table}[h!]
    \centering
    \small
    \begin{tabular}{|c|c|c|}
        \hline
        \textbf{Model} & \textbf{CSS PCP} & \textbf{CFS PCP} \\
        \hline
        No Context & 70.71\% $\pm$ 2.61 & 81.67\% $\pm$ 2.89 \\
        Text-Concat & 70.80\% $\pm$ 3.61 & 82.33\% $\pm$ 1.88 \\
        Embed-Concat & 70.97\% $\pm$ 1.19 & 83.00\% $\pm$ 1.98 \\
        GAT 3L (\textit{ours}) & 74.07\% $\pm$ 1.12 & 84.21\% $\pm$ 2.14  \\
        \hline
    \end{tabular}
    \caption{Percentage of Correct Predictions (PCP) for different models on Context-Sensitive Samples (CSS) and Context-Free Samples (CFS). Results show the mean percentage of correct predictions ($\pm$ 95\% CI) across all predictions, averaged over 10 model runs.}
    \label{tab:context_errors}
\end{table}

Table~\ref{tab:context_errors} shows that predicting Context-Sensitive Samples (CSS) is more challenging than Context-Free Samples (CFS) across all models. Our GAT model achieves the highest accuracy in both CSS and CFS settings, demonstrating the effectiveness of modeling conversational structure instead of treating context as a simple sequential input. Notably, the performance improvement of our GAT model is more pronounced for CSS cases. Compared to the No Context model, our GAT model achieves an improvement of 4.75\% in CSS, while the improvement in CFS is 3.11\%. Similarly, compared to the Text-Concat model, our GAT model outperforms by 4.62\% in CSS and 2.28\% in CFS. This relative over-performance illustrates that our model particularly enhances predictions in cases where understanding conversational context is essential.

\subsection{Case Analysis}
\label{subsec:case-analysis}

\begin{figure}[t]
  \includegraphics[width=\columnwidth]{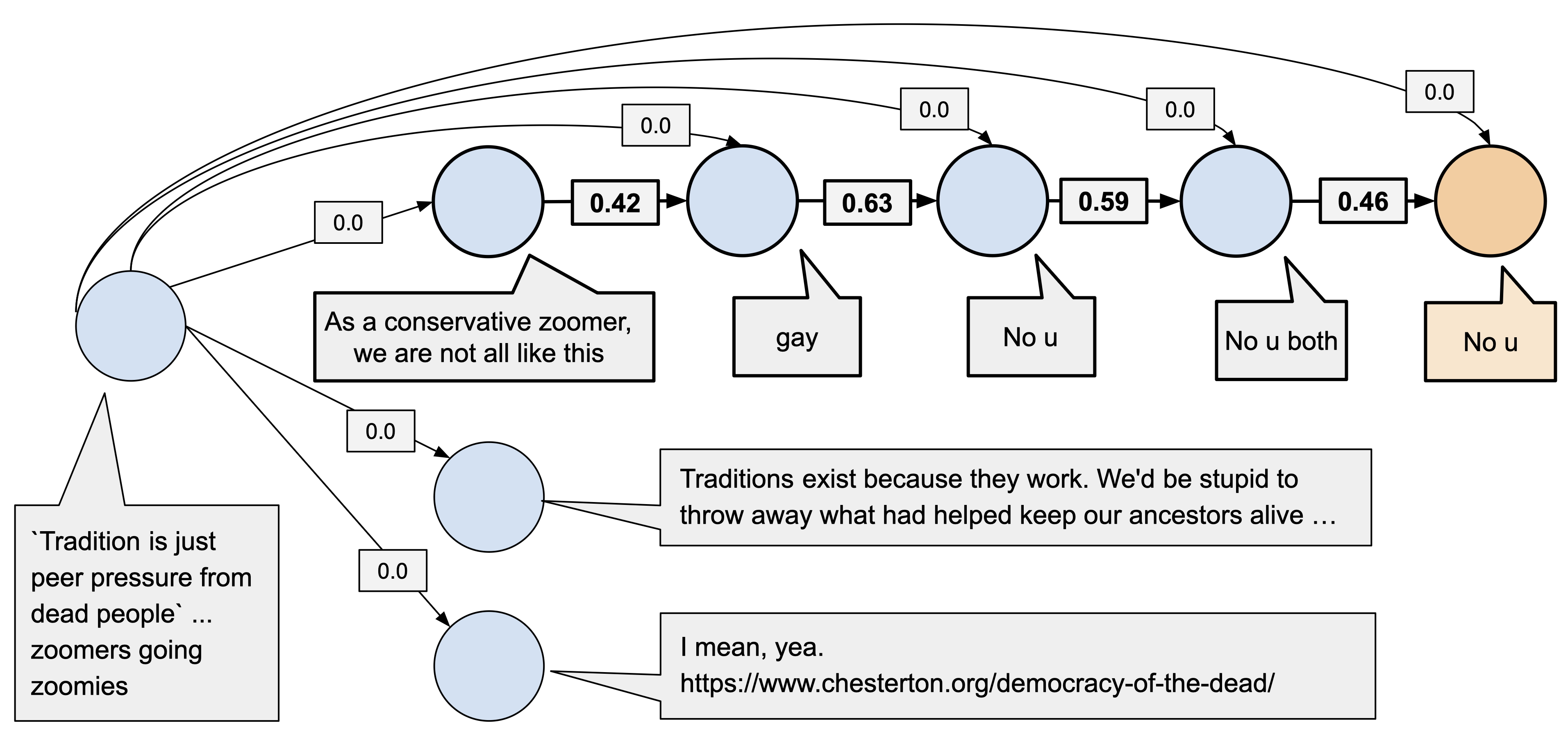}
  \caption{Example conversation graph with learned attention weights from the third layer of the best-performing GAT model. For readability, self-loop edges are omitted; their attention weights are one minus the sum of incoming edge weights.}
  \label{fig:attw_graph}
\end{figure}

To better understand our model’s behavior, we analyze a representative case from the test set—the conversation graph introduced in Section~\ref{sec:introduction}. This instance was incorrectly classified by the \textit{No Context} baseline but correctly identified as abusive by our graph-based model. The key contextual cue—a homophobic insult—appears three hops away from the target comment, a pattern common in large, multi-user conversations. As shown in Figure~\ref{fig:example}, abusive comments often trigger reactive responses, creating a snowball effect that requires tracing further up the conversation thread for proper interpretation.

To assess the impact of context depth, we conduct inference using the best-performing GAT model across 10 runs, varying the number of GAT layers from 1 to 5. As expected, models with only 1 or 2 layers fail to classify the target comment as abusive, whereas models with 3 or more layers succeed. This finding supports our hypothesis that short-context models lack the depth required to disambiguate meaning in threaded discussions, particularly in cases of abuse propagation where users reinforce or react to prior offensive content. Our results highlight the importance of sufficiently deep graph models (at least 3 GAT layers) for detecting abusive speech in complex, multi-user discussions such as those on Reddit.

To further interpret the model's predictions, we examine the learned attention weights of the \textit{GAT (3-Layers)} model at inference time (Figure~\ref{fig:attw_graph}). The model effectively assigns attention to relevant contextual nodes within the conversation graph. Notably, edges connecting the original post to other comments receive minimal attention, indicating that the post content does not contribute to determining the abusiveness of the target comment. In contrast, edges along the reply chain leading to the target comment receive higher attention weights, highlighting their importance in contextualizing abusiveness. This analysis demonstrates that the GAT layer dynamically identifies and prioritizes key conversational cues, reinforcing the effectiveness of structured context modeling in ALD.

\section{Conclusion and Future Work}
\label{sec:conclusion}

This study presents a novel methodology into integrating conversational context into ALD on Reddit threads. Unlike prior works that rely on limited context (e.g., only the previous comment or the initial post), we adopt a broader definition, incorporating all preceding comments. This approach is particularly suited for platforms features large, multi-user discussions where references to distant comments are common. To efficiently model extended context, we construct affordance-based conversation graphs, retaining only comments visible to users at the time of writing, based on Reddit’s UI rendering algorithm. This method ensures computational efficiency while preserving key contextual information, outperforming alternative graph constructions for ALD. 

Comparing our graph-based approach with conventional NLP models using flattened context, we demonstrate that explicitly modeling the topology of conversations significantly improves classification performance. Consistent with prior research, we find that simple context concatenation often fails to enhance ALD and can even degrade performance compared to context-agnostic models. Our analysis further indicates that our GAT model achieve greater performance gains in cases where context is crucial for disambiguation, highlighting the advantages of structured context modeling.

Overall, our findings highlight the advantages of GNNs over flat context models when combined with efficient graph construction and optimal context window selection. Our approach remains lightweight and scalable, making it well-suited for large-scale deployment on social media platforms.  

The review of existing datasets highlights a significant lack of publicly available ALD data that fully captures threaded conversations. Future research should focus on developing such datasets to facilitate cross-platform, multilingual, and large-scale evaluations of our approach. Additionally, we aim to enhance our model’s ability to detect implicit abuse and sarcasm while exploring multimodal extensions by integrating visual context—such as images in initial posts—which often provide crucial information for understanding conversations. Further, incorporating richer contextual signals, including user behaviors and social dynamics \citep{vidgen2019challenges, castelle2018linguistic}, remains an important direction. A promising avenue is the use of heterogeneous graph representations that fuse user embeddings, social interactions, and external media (e.g., images or videos) to more effectively model the complexities of online discourse.

\section*{Limitations}
\label{sec:limitations}

While this work highlights the effectiveness of graph-based methods for incorporating conversational context into Abusive Language Detection (ALD), it also exposes several limitations inherent to both our approach and the broader ALD research landscape.

\paragraph{Defining Abuse, Subjectivity and Biases} Abusive language detection lacks a universal definition, with studies adopting varying taxonomies for hate speech, toxicity, and offensiveness, leading to inconsistencies in annotation and evaluation \citep{vidgen2019challenges, fortuna2020toxic}. Existing datasets, such as CAD, reflect cultural and social biases, which can impact model predictions. Especially, CAD has been annotated by 12 annotators, mostly British English speakers, limiting generalizability. Additionally, Reddit-specific data with distinct language norms annotated by academic researchers introduces biases that can distort model predictions. For example, African American English (AAE) markers are often misclassified as abusive due to annotator bias \citep{sap2019risk}. \citet{vidgen2021introducing} attempted to improve annotation quality and consistency through a consensus-based approach, but the lack of access to initial annotator disagreements prevents deeper analysis of subjectivity. Such methods include integrating multi-annotator models \citep{davani2022dealing} and techniques addressing subjective annotation uncertainty \citep{rizos2020average, helwe2023MAFALDA} to enhance fairness.
\paragraph{Scalability and Computational Efficiency} While our approach demonstrates clear improvements over context-agnostic baselines, deploying such models on large-scale social media data requires optimizations to ensure efficiency without compromising performance. Our work introduced Affordance-Based pruning technics to reduce the conversation graph size while focusing on relevant context. However, using graph networks still adds costs, energy consumption, and computational overhead which should be considered when scaling to real-time applications.

In summary, while graph-based methods advance ALD, challenges remain in mitigating biases, enriching contextual modeling, ensuring cross-platform generalizability, and improving scalability. Addressing these will be crucial for fair, efficient, and practical ALD systems.

\section*{Ethical Considerations}
\label{sec:ethics}

\paragraph{Potential Risks}  
Our work contributes to the development of context-aware models for abusive language detection (ALD), which can aid in moderating harmful content on social media. However, automatic ALD systems present inherent risks, particularly when deployed without human oversight. False positives may result in the unjust removal or suppression of benign content, potentially restricting freedom of expression, while false negatives may fail to detect harmful speech, enabling the spread of abuse. Given these limitations, human oversight is essential, and users should retain the right to appeal algorithmic moderation decisions. Future work should focus on improving robustness, reducing errors, and mitigating biases to enhance the reliability of ALD systems. Additionally, the use of ALD models must align with ethical guidelines and platform policies to prevent misuse. Potential risks include weaponization for mass reporting, over-censorship, or the reinforcement of societal biases. 
\paragraph{Data Privacy and Bias}  
All experiments were conducted using the Contextual Abuse Dataset (CAD) \citep{vidgen2021introducing}, a publicly available dataset derived from Reddit. The dataset has been anonymized to remove personally identifiable information. However, ALD models trained on existing datasets may inherit biases from annotation processes, as discussed in Section~\ref{sec:limitations}. Biases related to cultural context, dialects (e.g., African American English), or platform-specific discourse norms can lead to disproportionate misclassifications. Addressing these biases requires ongoing evaluation, diverse datasets, and improvements in annotation methodologies to mitigate unintended harms.  
\paragraph{Transparency and Reproducibility}  
We provide a detailed account of our methodology, dataset statistics, and hyperparameter settings to facilitate transparency and reproducibility. The code is released publicly to encourage further research and independent evaluations.

\bibliography{acl_latex}

\begin{thebibliography}{48}
\providecommand{\natexlab}[1]{#1}

\bibitem[{Agarwal et~al.(2023)Agarwal, Young, Joglekar, and Sastry}]{agarwal2023graph}
Vibhor Agarwal, Anthony~P. Young, Sagar Joglekar, and Nishanth Sastry. 2023.
\newblock \href {https://doi.org/10.1145/3624579} {A graph-based context-aware model to understand online conversations}.
\newblock \emph{ACM Transactions on the Web}, pages 1--25.

\bibitem[{AI(2024)}]{deepseekv3}
DeepSeek AI. 2024.
\newblock \href {https://arxiv.org/html/2412.19437v1} {Deepseek-v3: Advancing scalable mixture-of-experts language models}.
\newblock \emph{arXiv preprint}, arXiv:2412.19437.

\bibitem[{{Arthur Heitmann, Stas Bekman}()}]{arcticshift2024}
{Arthur Heitmann, Stas Bekman}.
\newblock Arctic shift download tool.
\newblock \url{https://arctic-shift.photon-reddit.com/download-tool}.
\newblock Accessed: Aug. 30, 2024.

\bibitem[{Beatty(2020)}]{beatty2020graph}
Matthew Beatty. 2020.
\newblock \href {https://doi.org/10.1109/ASONAM49781.2020.9381473} {Graph-based methods to detect hate speech diffusion on twitter}.
\newblock In \emph{Proceedings of the 2020 IEEE/ACM International Conference on Advances in Social Networks Analysis and Mining (ASONAM)}, pages 502--506, The Hague, Netherlands. IEEE.

\bibitem[{Bourgeade et~al.(2023)Bourgeade, Chiril, Benamara, and Moriceau}]{bourgeade2023what}
Tom Bourgeade, Patricia Chiril, Farah Benamara, and Véronique Moriceau. 2023.
\newblock \href {https://doi.org/10.18653/v1/2023.eacl-main.254} {What did you learn to hate? a topic-oriented analysis of generalization in hate speech detection}.
\newblock In \emph{Proceedings of the 17th Conference of the European Chapter of the Association for Computational Linguistics}, pages 3495--3508, Dubrovnik, Croatia. Association for Computational Linguistics.

\bibitem[{Bourgeade et~al.(2024)Bourgeade, Li, Benamara, Moriceau, Su, and Sun}]{bourgeade2024humans}
Tom Bourgeade, Zongmin Li, Farah Benamara, V{\'e}ronique Moriceau, Jian Su, and Aixin Sun. 2024.
\newblock \href {https://aclanthology.org/2024.lrec-main.740} {Humans need context, what about machines? investigating conversational context in abusive language detection}.
\newblock In \emph{Proceedings of the 2024 Joint International Conference on Computational Linguistics, Language Resources and Evaluation (LREC-COLING 2024)}, pages 8438--8452, Torino, Italia. ELRA and ICCL.

\bibitem[{Brown et~al.(2020)Brown, Mann, Ryder, Subbiah, Kaplan, Dhariwal, Neelakantan, Shyam, Sastry, Askell, Agarwal, Herbert{-}Voss, Krueger, Henighan, Child, Ramesh, Ziegler, Wu, Winter, Hesse, Chen, Sigler, Litwin, Gray, Chess, Clark, Berner, McCandlish, Radford, Sutskever, and Amodei}]{brown2020language}
Tom~B. Brown, Benjamin Mann, Nick Ryder, Melanie Subbiah, Jared Kaplan, Prafulla Dhariwal, Arvind Neelakantan, Pranav Shyam, Girish Sastry, Amanda Askell, Sandhini Agarwal, Ariel Herbert{-}Voss, Gretchen Krueger, Tom Henighan, Rewon Child, Aditya Ramesh, Daniel~M. Ziegler, Jeffrey Wu, Clemens Winter, Christopher Hesse, Mark Chen, Eric Sigler, Mateusz Litwin, Scott Gray, Benjamin Chess, Jack Clark, Christopher Berner, Sam McCandlish, Alec Radford, Ilya Sutskever, and Dario Amodei. 2020.
\newblock \href {https://arxiv.org/abs/2005.14165} {Language models are few-shot learners}.
\newblock \emph{CoRR}, abs/2005.14165.

\bibitem[{Castelle(2018)}]{castelle2018linguistic}
Michael Castelle. 2018.
\newblock \href {https://doi.org/10.18653/v1/W18-5120} {The linguistic ideologies of deep abusive language classification}.
\newblock In \emph{Proceedings of the Workshop on Abusive Language Online (ALW)}, pages 160--170, Melbourne, Australia. Association for Computational Linguistics.

\bibitem[{Cecillon et~al.(2021)Cecillon, Labatut, Dufour, and Linares}]{cecillon2021graph}
No{\'e} Cecillon, Vincent Labatut, Richard Dufour, and Georges Linares. 2021.
\newblock \href {https://doi.org/10.1007/s42979-020-00406-w} {Graph embeddings for abusive language detection}.
\newblock \emph{SN Computer Science}, 2(1):1--15.
\newblock LIA (Laboratoire Informatique d'Avignon).

\bibitem[{Chiu and Alexander(2021)}]{chiu2021detecting}
Ke{-}Li Chiu and Rohan Alexander. 2021.
\newblock \href {https://arxiv.org/abs/2103.12407} {Detecting hate speech with {GPT-3}}.
\newblock \emph{CoRR}, abs/2103.12407.

\bibitem[{Davani et~al.(2022)Davani, Díaz, and Prabhakaran}]{davani2022dealing}
Aida~Mostafazadeh Davani, Mark Díaz, and Vinodkumar Prabhakaran. 2022.
\newblock \href {https://doi.org/10.1162/tacl_a_00449} {Dealing with disagreements: Looking beyond the majority vote in subjective annotations}.
\newblock \emph{Transactions of the Association for Computational Linguistics}, 10:92--110.

\bibitem[{Davidson et~al.(2017)Davidson, Warmsley, Macy, and Weber}]{davidson2017automated}
Thomas Davidson, Dana Warmsley, Michael Macy, and Ingmar Weber. 2017.
\newblock \href {https://doi.org/10.1609/icwsm.v11i1.14955} {Automated hate speech detection and the problem of offensive language}.
\newblock In \emph{Proceedings of the International AAAI Conference on Web and Social Media (ICWSM)}, volume~11, pages 512--515.

\bibitem[{DeepSeek-AI et~al.(2024)DeepSeek-AI, Liu, Feng, Wang, Wang, Liu, Zhao, Deng, Ruan, Dai, Guo, Yang, Chen, Ji, Li, Lin, Luo, Hao, Chen, Li, Zhang, Xu, Yang, Zhang, Ding, Xin, Gao, Li, Qu, Cai, Liang, Guo, Ni, Li, Chen, Yuan, Qiu, Song, Dong, Gao, Guan, Wang, Zhang, Xu, Xia, Zhao, Zhang, Li, Wang, Zhang, Zhang, Tang, Li, Tian, Huang, Wang, Zhang, Zhu, Chen, Du, Chen, Jin, Ge, Pan, Xu, Chen, Li, Lu, Zhou, Chen, Wu, Ye, Ma, Wang, Zhou, Yu, Zhou, Zheng, Wang, Pei, Yuan, Sun, Xiao, Zeng, An, Liu, Liang, Gao, Zhang, Li, Jin, Wang, Bi, Liu, Wang, Shen, Chen, Chen, Nie, Sun et~al.}]{deepseekv2}
DeepSeek-AI, Aixin Liu, Bei Feng, Bin Wang, Bingxuan Wang, Bo~Liu, Chenggang Zhao, Chengqi Deng, Chong Ruan, Damai Dai, Daya Guo, Dejian Yang, Deli Chen, Dongjie Ji, Erhang Li, Fangyun Lin, Fuli Luo, Guangbo Hao, Guanting Chen, Guowei Li, H.~Zhang, Hanwei Xu, Hao Yang, Haowei Zhang, Honghui Ding, Huajian Xin, Huazuo Gao, Hui Li, Hui Qu, J.L. Cai, Jian Liang, Jianzhong Guo, Jiaqi Ni, Jiashi Li, Jin Chen, Jingyang Yuan, Junjie Qiu, Junxiao Song, Kai Dong, Kaige Gao, Kang Guan, Lean Wang, Lecong Zhang, Lei Xu, Leyi Xia, Liang Zhao, Liyue Zhang, Meng Li, Miaojun Wang, Mingchuan Zhang, Minghua Zhang, Minghui Tang, Mingming Li, Ning Tian, Panpan Huang, Peiyi Wang, Peng Zhang, Qihao Zhu, Qinyu Chen, Qiushi Du, R.J. Chen, R.L. Jin, Ruiqi Ge, Ruizhe Pan, Runxin Xu, Ruyi Chen, S.S. Li, Shanghao Lu, Shangyan Zhou, Shanhuang Chen, Shaoqing Wu, Shengfeng Ye, Shirong Ma, Shiyu Wang, Shuang Zhou, Shuiping Yu, Shunfeng Zhou, Size Zheng, T.~Wang, Tian Pei, Tian Yuan, Tianyu Sun, W.L. Xiao, Wangding Zeng, Wei An, Wen Liu,
  Wenfeng Liang, Wenjun Gao, Wentao Zhang, X.Q. Li, Xiangyue Jin, Xianzu Wang, Xiao Bi, Xiaodong Liu, Xiaohan Wang, Xiaojin Shen, Xiaokang Chen, Xiaosha Chen, Xiaotao Nie, Xiaowen Sun, et~al. 2024.
\newblock \href {https://arxiv.org/abs/2405.04434} {Deepseek-v2: A strong, economical, and efficient mixture-of-experts language model}.
\newblock \emph{arXiv preprint}, arXiv:2405.04434.

\bibitem[{Devlin et~al.(2019)Devlin, Chang, Lee, and Toutanova}]{devlin2019bert}
Jacob Devlin, Ming-Wei Chang, Kenton Lee, and Kristina Toutanova. 2019.
\newblock Bert: Pre-training of deep bidirectional transformers for language understanding.
\newblock In \emph{Proceedings of the 2019 Conference of the North American Chapter of the Association for Computational Linguistics: Human Language Technologies, Volume 1 (Long and Short Papers)}, pages 4171--4186, Minneapolis, Minnesota. Association for Computational Linguistics.

\bibitem[{Duggan(2017)}]{duggan2017onlineharassment}
Maeve Duggan. 2017.
\newblock \href {https://www.pewresearch.org/internet/2017/07/11/online-harassment-2017} {Online harassment 2017}.

\bibitem[{Duong et~al.(2022)Duong, Zhang, and Lu}]{duong2022hatenet}
Charles Duong, Lei Zhang, and Chang-Tien Lu. 2022.
\newblock \href {https://doi.org/10.1109/BigData55660.2022.10020510} {Hatenet: A graph convolutional network approach to hate speech detection}.
\newblock In \emph{2022 IEEE International Conference on Big Data (Big Data)}, page 5698–5707.

\bibitem[{Fey and Lenssen(2019)}]{fey2019fast}
Matthias Fey and Jan~E. Lenssen. 2019.
\newblock Fast graph representation learning with {PyTorch Geometric}.
\newblock In \emph{ICLR Workshop on Representation Learning on Graphs and Manifolds}.

\bibitem[{Fortuna et~al.(2020)Fortuna, Soler, and Wanner}]{fortuna2020toxic}
Paula Fortuna, Juan Soler, and Leo Wanner. 2020.
\newblock Toxic, hateful, offensive or abusive? what are we really classifying? an empirical analysis of hate speech datasets.
\newblock In \emph{Proceedings of the Twelfth Language Resources and Evaluation Conference}, pages 6786--6794, Marseille, France. European Language Resources Association.
\newblock Accessed: Jul. 30, 2024. [Online]. Available: https://aclanthology.org/2020.lrec-1.838.

\bibitem[{Founta et~al.(2018)Founta, Djouvas, Chatzakou, Leontiadis, Blackburn, Stringhini, Vakali, Sirivianos, and Kourtellis}]{founta2018large}
Antigoni Founta, Constantinos Djouvas, Despoina Chatzakou, Ilias Leontiadis, Jeremy Blackburn, Gianluca Stringhini, Athena Vakali, Michael Sirivianos, and Nicolas Kourtellis. 2018.
\newblock \href {https://doi.org/10.1609/icwsm.v12i1.14991} {Large scale crowdsourcing and characterization of twitter abusive behavior}.
\newblock \emph{Proceedings of the International AAAI Conference on Web and Social Media}, 12(1).

\bibitem[{Golbeck et~al.(2017)Golbeck, Gnanasekaran, Gunasekaran, Hoffman, Hottle, Jienjitlert, Khare, Lau, Martindale, Naik, Nixon, Ashktorab, Ramachandran, Rogers, Rogers, Sarin, Shahane, Thanki, Vengataraman, Wan, Wu, Banjo, Berlinger, Bhagwan, Buntain, Cheakalos, Geller, and Gregory}]{golbeck2017large}
Jennifer Golbeck, Rajesh~Kumar Gnanasekaran, Raja~Rajan Gunasekaran, Kelly~M. Hoffman, Jenny Hottle, Vichita Jienjitlert, Shivika Khare, Ryan Lau, Marianna~J. Martindale, Shalmali Naik, Heather~L. Nixon, Zahra Ashktorab, Piyush Ramachandran, Kristine~M. Rogers, Lisa Rogers, Meghna~Sardana Sarin, Gaurav Shahane, Jayanee Thanki, Priyanka Vengataraman, Zijian Wan, Derek~Michael Wu, Rashad~O. Banjo, Alexandra Berlinger, Siddharth Bhagwan, Cody Buntain, Paul Cheakalos, Alicia~A. Geller, and Quint Gregory. 2017.
\newblock \href {https://doi.org/10.1145/3091478.3091509} {A large labeled corpus for online harassment research}.
\newblock In \emph{Proceedings of the 2017 ACM on Web Science Conference (WebSci '17)}, pages 229--233. ACM Press.

\bibitem[{Guo et~al.(2024)Guo, Hu, Mu, Shi, Zhao, Vishwamitra, and Hu}]{guo2024investigation}
Keyan Guo, Alexander Hu, Jaden Mu, Ziheng Shi, Ziming Zhao, Nishant Vishwamitra, and Hongxin Hu. 2024.
\newblock \href {https://arxiv.org/abs/2401.03346} {An investigation of large language models for real-world hate speech detection}.
\newblock \emph{Preprint}, arXiv:2401.03346.

\bibitem[{Hebert et~al.(2022)Hebert, Golab, and Cohen}]{hebert2022predicting}
Liam Hebert, Lukasz Golab, and Robin Cohen. 2022.
\newblock \href {https://doi.org/10.1109/WI-IAT55865.2022.00012} {Predicting hateful discussions on reddit using graph transformer networks and communal context}.
\newblock In \emph{2022 IEEE/WIC/ACM International Joint Conference on Web Intelligence and Intelligent Agent Technology (WI-IAT)}, pages 9--17.

\bibitem[{Hebert et~al.(2024)Hebert, Sahu, Guo, Sreenivas, Golab, and Cohen}]{hebert2024multi}
Liam Hebert, Gaurav Sahu, Yuxuan Guo, Nanda~Kishore Sreenivas, Lukasz Golab, and Robin Cohen. 2024.
\newblock \href {https://doi.org/10.1609/aaai.v38i20.30213} {Multi-modal discussion transformer: Integrating text, images and graph transformers to detect hate speech on social media}.
\newblock \emph{Proceedings of the AAAI Conference on Artificial Intelligence}, 38:20.

\bibitem[{Helwe et~al.(2023)Helwe, Calamai, Paris, Clavel, and Suchanek}]{helwe2023MAFALDA}
Chadi Helwe, Tom Calamai, Pierre-Henri Paris, Chloé Clavel, and Fabian Suchanek. 2023.
\newblock \href {https://arxiv.org/abs/2311.09761} {{MAFALDA}: A benchmark and comprehensive study of fallacy detection and classification}.
\newblock \emph{arXiv preprint arXiv:2311.09761}.

\bibitem[{{\.I}htiyar et~al.(2023){\.I}htiyar, {\"O}zdemir, Ereng{\"u}l, and {\"O}zg{\"u}r}]{ihtiyar2023dataset}
Musa {\.I}htiyar, {\"O}mer {\"O}zdemir, Mustafa Ereng{\"u}l, and Arzucan {\"O}zg{\"u}r. 2023.
\newblock \href {https://doi.org/10.18653/v1/2023.findings-emnlp.103} {A dataset for investigating the impact of context for offensive language detection in tweets}.
\newblock In \emph{Findings of the Association for Computational Linguistics: EMNLP 2023}, pages 1543--1549, Singapore. Association for Computational Linguistics.

\bibitem[{Ive et~al.(2021)Ive, Anuchitanukul, and Specia}]{ive2021revisiting}
Julia Ive, Atijit Anuchitanukul, and Lucia Specia. 2021.
\newblock \href {https://arxiv.org/abs/2111.12447} {Revisiting contextual toxicity detection in conversations}.
\newblock \emph{CoRR}, abs/2111.12447.

\bibitem[{Loshchilov and Hutter(2019)}]{loshchilov2019decoupled}
Ilya Loshchilov and Frank Hutter. 2019.
\newblock Decoupled weight decay regularization.
\newblock \emph{International Conference on Learning Representations (ICLR)}.

\bibitem[{Meng et~al.(2023)Meng, Suresh, Lee, and Chakraborty}]{meng2023predicting}
Qing Meng, Tharun Suresh, Roy Ka-Wei Lee, and Tanmoy Chakraborty. 2023.
\newblock \href {https://doi.org/10.48550/arXiv.2206.08406} {Predicting hate intensity of twitter conversation threads}.
\newblock \emph{arXiv}, (arXiv:2206.08406).
\newblock ArXiv:2206.08406 [cs].

\bibitem[{Menini et~al.(2021)Menini, Aprosio, and Tonelli}]{menini2021abuse}
Stefano Menini, Alessio~Palmero Aprosio, and Sara Tonelli. 2021.
\newblock \href {https://arxiv.org/abs/2103.14916} {Abuse is contextual, what about nlp? the role of context in abusive language annotation and detection}.
\newblock \emph{CoRR}, abs/2103.14916.

\bibitem[{Papegnies et~al.(2019)Papegnies, Labatut, Dufour, and Linares}]{papegnies2019conversational}
Etienne Papegnies, Vincent Labatut, Richard Dufour, and Georges Linares. 2019.
\newblock \href {https://doi.org/10.1109/TCSS.2018.2887240} {Conversational networks for automatic online moderation}.
\newblock \emph{IEEE Transactions on Computational Social Systems}.
\newblock Also available as arXiv preprint: arXiv:1901.11281.

\bibitem[{Paszke et~al.(2019)Paszke, Gross, Massa, Lerer, Bradbury, Chanan, Killeen, Lin, Gimelshein, Antiga, Desmaison, Kopf, Yang, DeVito, Raison, Tejani, Chilamkurthy, Steiner, Fang, Bai, and Chintala}]{paszke2019pytorch}
Adam Paszke, Sam Gross, Francisco Massa, Adam Lerer, James Bradbury, Gregory Chanan, Trevor Killeen, Zeming Lin, Natalia Gimelshein, Luca Antiga, Alban Desmaison, Andreas Kopf, Edward Yang, Zachary DeVito, Martin Raison, Alykhan Tejani, Sasank Chilamkurthy, Benoit Steiner, Lu~Fang, Junjie Bai, and Soumith Chintala. 2019.
\newblock Pytorch: An imperative style, high-performance deep learning library.
\newblock In \emph{Advances in Neural Information Processing Systems (NeurIPS)}.

\bibitem[{Pavlopoulos et~al.(2020)Pavlopoulos, Sorensen, Dixon, Thain, and Androutsopoulos}]{pavlopoulos2020toxicity}
John Pavlopoulos, Jeffrey Sorensen, Lucas Dixon, Nithum Thain, and Ion Androutsopoulos. 2020.
\newblock \href {https://doi.org/10.18653/v1/2020.acl-main.396} {Toxicity detection: Does context really matter?}
\newblock In \emph{Proceedings of the 58th Annual Meeting of the Association for Computational Linguistics}, page 4296–4305, Online. Association for Computational Linguistics.

\bibitem[{Qian et~al.(2019)Qian, Bethke, Liu, Belding, and Wang}]{qian2019benchmark}
Jing Qian, Anna Bethke, Yinyin Liu, Elizabeth Belding, and William~Yang Wang. 2019.
\newblock \href {https://doi.org/10.48550/arXiv.1909.04251} {A benchmark dataset for learning to intervene in online hate speech}.
\newblock \emph{arXiv}.

\bibitem[{Rizos and Schuller(2020)}]{rizos2020average}
Georgios Rizos and Björn~W. Schuller. 2020.
\newblock \href {https://doi.org/10.1007/978-3-030-50146-4_4} {Average jane, where art thou? – recent avenues in efficient machine learning under subjectivity uncertainty}.
\newblock In \emph{Proceedings of the 18th International Conference on Information Processing and Management of Uncertainty in Knowledge-Based Systems (IPMU 2020)}, pages 42--55. Springer, Cham.

\bibitem[{Röttger et~al.(2022)Röttger, Vidgen, Hovy, and Pierrehumbert}]{rottger2022contrasting}
Paul Röttger, Bertie Vidgen, Dirk Hovy, and Janet Pierrehumbert. 2022.
\newblock \href {https://doi.org/10.18653/v1/2022.naacl-main.13} {Two contrasting data annotation paradigms for subjective nlp tasks}.
\newblock In \emph{Proceedings of the 2022 Conference of the North American Chapter of the Association for Computational Linguistics: Human Language Technologies}, pages 175--190, Seattle, United States. Association for Computational Linguistics.

\bibitem[{Sap et~al.(2019)Sap, Card, Gabriel, Choi, and Smith}]{sap2019risk}
Maarten Sap, Dallas Card, Saadia Gabriel, Yejin Choi, and Noah~A. Smith. 2019.
\newblock \href {https://doi.org/10.18653/v1/P19-1163} {The risk of racial bias in hate speech detection}.
\newblock In \emph{Proceedings of the 57th Annual Meeting of the Association for Computational Linguistics}, pages 1668--1678, Florence, Italy. Association for Computational Linguistics.

\bibitem[{Saveski et~al.(2021)Saveski, Roy, and Roy}]{saveski2021structure}
Martin Saveski, Brandon Roy, and Deb Roy. 2021.
\newblock \href {https://doi.org/10.1145/3442381.3449861} {The structure of toxic conversations on twitter}.
\newblock In \emph{Proceedings of the Web Conference 2021}, WWW ’21, page 1086–1097, New York, NY, USA. Association for Computing Machinery.

\bibitem[{Touvron et~al.(2023)Touvron, Martin, Stone, Albert, Almahairi, Benajiba, Caudwell et~al.}]{touvron2023llama}
Hugo Touvron, Louis Martin, Kevin Stone, Peter Albert, Amjad Almahairi, Yasmine Benajiba, Rene Caudwell, et~al. 2023.
\newblock \href {https://arxiv.org/pdf/2307.09288} {Llama 2: Open foundation and fine-tuned chat models}.
\newblock \emph{arXiv preprint}, arXiv:2307.09288.

\bibitem[{Velickovic et~al.(2018)Velickovic, Cucurull, Casanova, Romero, Lio, and Bengio}]{velickovic2018graph}
Petar Velickovic, Guillem Cucurull, Arantxa Casanova, Adriana Romero, Pietro Lio, and Yoshua Bengio. 2018.
\newblock Graph attention networks.
\newblock In \emph{International Conference on Learning Representations (ICLR)}.

\bibitem[{Vidgen et~al.(2019)Vidgen, Harris, Nguyen, Tromble, Hale, and Margetts}]{vidgen2019challenges}
Bertie Vidgen, Alex Harris, Dong Nguyen, Rebekah Tromble, Scott Hale, and Helen Margetts. 2019.
\newblock \href {https://doi.org/10.18653/v1/W19-3509} {Challenges and frontiers in abusive content detection}.
\newblock In \emph{Proceedings of the Third Workshop on Abusive Language Online}, pages 80--93, Florence, Italy. Association for Computational Linguistics.

\bibitem[{Vidgen et~al.(2021)Vidgen, Nguyen, Margetts, Rossini, and Tromble}]{vidgen2021introducing}
Bertie Vidgen, Dong Nguyen, Helen Margetts, Patricia Rossini, and Rebekah Tromble. 2021.
\newblock \href {https://doi.org/10.18653/v1/2021.naacl-main.182} {Introducing cad: the contextual abuse dataset}.
\newblock In \emph{Proceedings of the 2021 Conference of the North American Chapter of the Association for Computational Linguistics: Human Language Technologies}, pages 2289--2303, Online. Association for Computational Linguistics.

\bibitem[{Wang et~al.(2020)Wang, Fu, and Lu}]{wang2020sosnet}
Jason Wang, Kaiqun Fu, and Chang-Tien Lu. 2020.
\newblock \href {https://people.cs.vt.edu/ctlu/Publication/2020/IEEE-BD-SOSNet-Wang.pdf} {Sosnet: A graph convolutional network approach to fine-grained cyberbullying detection}.
\newblock In \emph{2020 IEEE International Conference on Big Data (BigData)}, pages 1699--1708. IEEE.

\bibitem[{Waseem and Hovy(2016)}]{waseem2016hateful}
Zeerak Waseem and Dirk Hovy. 2016.
\newblock \href {https://doi.org/10.18653/v1/N16-2013} {Hateful symbols or hateful people? predictive features for hate speech detection on {T}witter}.
\newblock In \emph{Proceedings of the {NAACL} Student Research Workshop}, pages 88--93, San Diego, California. Association for Computational Linguistics.

\bibitem[{Wei et~al.(2022)Wei, Wang, Schuurmans, Bosma, Chi, Le, and Zhou}]{wei2022chain}
Jason Wei, Xuezhi Wang, Dale Schuurmans, Maarten Bosma, Ed~H. Chi, Quoc Le, and Denny Zhou. 2022.
\newblock \href {https://arxiv.org/abs/2201.11903} {Chain of thought prompting elicits reasoning in large language models}.
\newblock \emph{CoRR}, abs/2201.11903.

\bibitem[{Wolf et~al.(2020)Wolf, Debut, Sanh, Chaumond, Delangue, Moi, Cistac, Rault, Louf, Funtowicz, Davison, Shleifer, von Platen, Ma, Jernite, Plu, Xu, Scao, Gugger, Drame, Lhoest, and Rush}]{wolf2020transformers}
Thomas Wolf, Lysandre Debut, Victor Sanh, Julien Chaumond, Clement Delangue, Anthony Moi, Pierric Cistac, Tim Rault, Remi Louf, Morgan Funtowicz, Joe Davison, Sam Shleifer, Patrick von Platen, Clara Ma, Yacine Jernite, Julien Plu, Canwen Xu, Teven~Le Scao, Sylvain Gugger, Mariama Drame, Quentin Lhoest, and Alexander~M. Rush. 2020.
\newblock Transformers: State-of-the-art natural language processing.
\newblock In \emph{Proceedings of the 2020 Conference on Empirical Methods in Natural Language Processing: System Demonstrations}.

\bibitem[{Ying et~al.(2021)Ying, Cai, Luo, Zheng, Ke, He, Shen, and Liu}]{ying2021dotransformers}
Chengxuan Ying, Tianle Cai, Shengjie Luo, Shuxin Zheng, Guolin Ke, Di~He, Yanming Shen, and Tie-Yan Liu. 2021.
\newblock Do transformers really perform badly for graph representation?
\newblock \emph{Advances in Neural Information Processing Systems}, 34:28877--28888.

\bibitem[{Yu et~al.(2022)Yu, Blanco, and Hong}]{yu2022hate}
Xinchen Yu, Eduardo Blanco, and Lingzi Hong. 2022.
\newblock \href {https://doi.org/10.18653/v1/2022.naacl-main.433} {Hate speech and counter speech detection: Conversational context does matter}.
\newblock In \emph{Proceedings of the 2022 Conference of the North American Chapter of the Association for Computational Linguistics: Human Language Technologies}, pages 5918--5930, Seattle, United States. Association for Computational Linguistics.

\bibitem[{Zayats and Ostendorf(2018)}]{zayats2018conversation}
Victoria Zayats and Mari Ostendorf. 2018.
\newblock \href {https://doi.org/10.1162/tacl_a_00009} {Conversation modeling on reddit using a graph-structured lstm}.
\newblock \emph{Transactions of the Association for Computational Linguistics}, 6:121--132.

\end{thebibliography}

\appendix
\section{Appendix}
\label{sec:appendix}

\subsection{Text Embeddings}
\label{subsec:text-embeddings}

We evaluated multiple text encoders for generating text embeddings in our models. Table~\ref{tab:text-models} reports the F1-scores of \textit{No Context} classifiers fine-tuned on the CAD dataset using different base encoders. BERT achieved the highest performance and was selected for all experiments. Importantly, our results are independent of this choice, as the same encoder was used across all baselines and graph-based models to ensure fair comparison.  

\begin{table}[h!]
    \centering
    \small
    \begin{tabular}{|c|c|}
        \hline
        \textbf{Model} & \textbf{F1 Score} \\
        \hline
        BERT & 0.745 \\
        RoBERTa  & 0.633 \\
        XLM-R & 0.718 \\
        Modern-BERT & 0.667 \\
        \hline
    \end{tabular}
    \caption{F1-scores for \textit{No Context} models fine-tuned on the adapted CAD dataset.}
    \label{tab:text-models}
\end{table}

\subsection{Graph Construction Methods}
\label{subsec:graph-construction}

We tested various graph construction methods to determine the most effective approach for modeling conversational context. Directed graphs outperformed undirected ones, as they better capture reply relationships. Additionally, we experimented with temporal edges \citep{zayats2018conversation}, linking sibling comments chronologically to model discussion flow, but this did not improve performance and was excluded from the final model.  

\begin{figure*}[t]
    \includegraphics[width=0.31\linewidth]{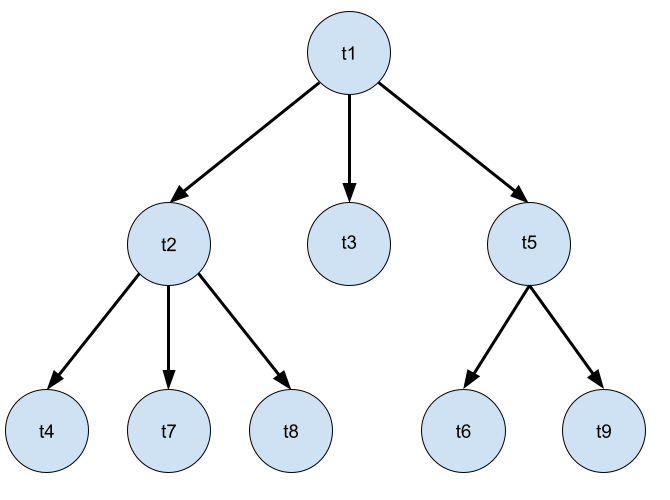} \hfill
    \includegraphics[width=0.31\linewidth]{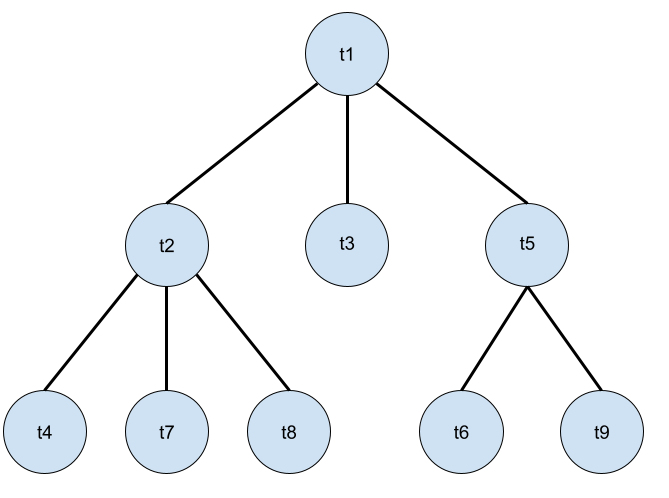} \hfill
    \includegraphics[width=0.31\linewidth]{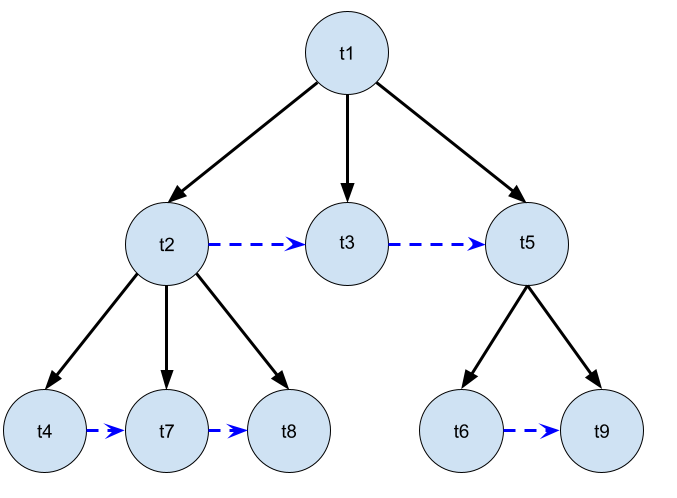} 
  \caption {Diagram of Reddit conversation graphs constructed using different edge methods. Node labels $t_i$ indicate comment publication times, with $t_i < t_j$ if $i < j$. From left to right: directed graph, undirected graph, and directed graph with temporal edges.}
\label{fig:graph-edge-construction}
\end{figure*}

Figure~\ref{fig:graph-edge-construction} illustrates the different graph structures, and Table~\ref{tab:graph-constr} reports their respective F1-scores. All methods were evaluated on the 3-layer GAT architecture (Section~\ref{subsec:model-archi}). The directed graph without temporal edges achieved the highest performance and was used in all experiments.  

\begin{table}[h!]
    \centering
    \small
    \begin{tabular}{|c|c|}
        \hline
        \textbf{Graph Type} & \textbf{F1 Score} \\
        \hline
        Directed & 0.765 \\
        Undirected  &  0.757 \\
        Directed + Temporal Edges & 0.761 \\
        Undirected + Temporal Edges & 0.756 \\
        \hline
    \end{tabular}
    \caption{F1-scores for different graph construction methods.}
    \label{tab:graph-constr}
\end{table}

\subsection{Trimming Strategies}
\label{subsec:trim-strat}

We evaluated different trimming strategies for constructing conversation graphs. Our primary approach follows an \textit{affordance-based} strategy, aligning with Reddit's UI rendering algorithm (Section~\ref{subsec:context-graph}). As an alternative, we tested a \textit{most recent} strategy, which removes all comments posted after the target comment and retains the 25 most recent preceding comments, matching the maximum node count in affordance-based graphs.  

\begin{figure*}[t]
  \includegraphics[width=0.48\linewidth]{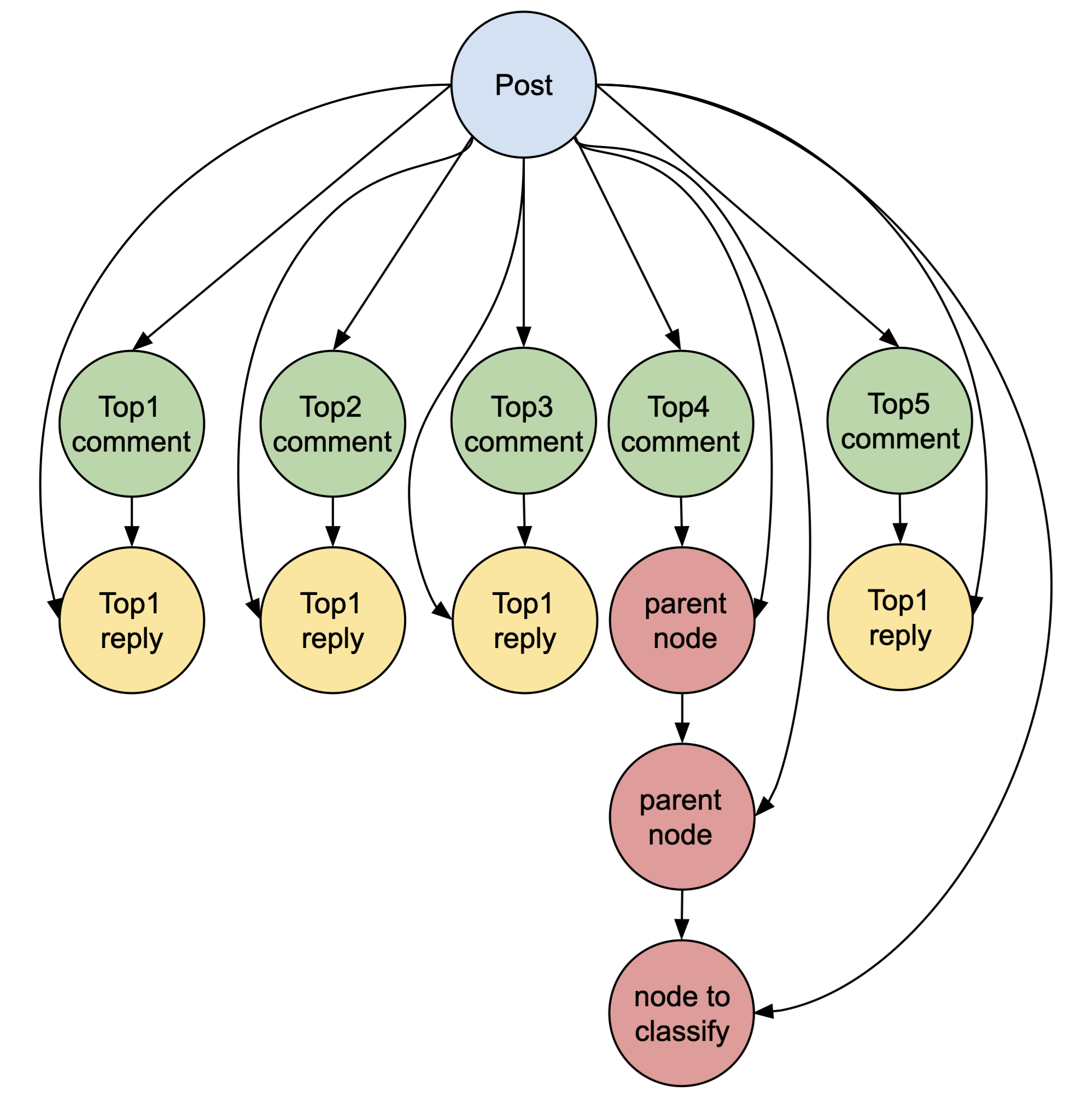} \hfill
  \includegraphics[width=0.48\linewidth]{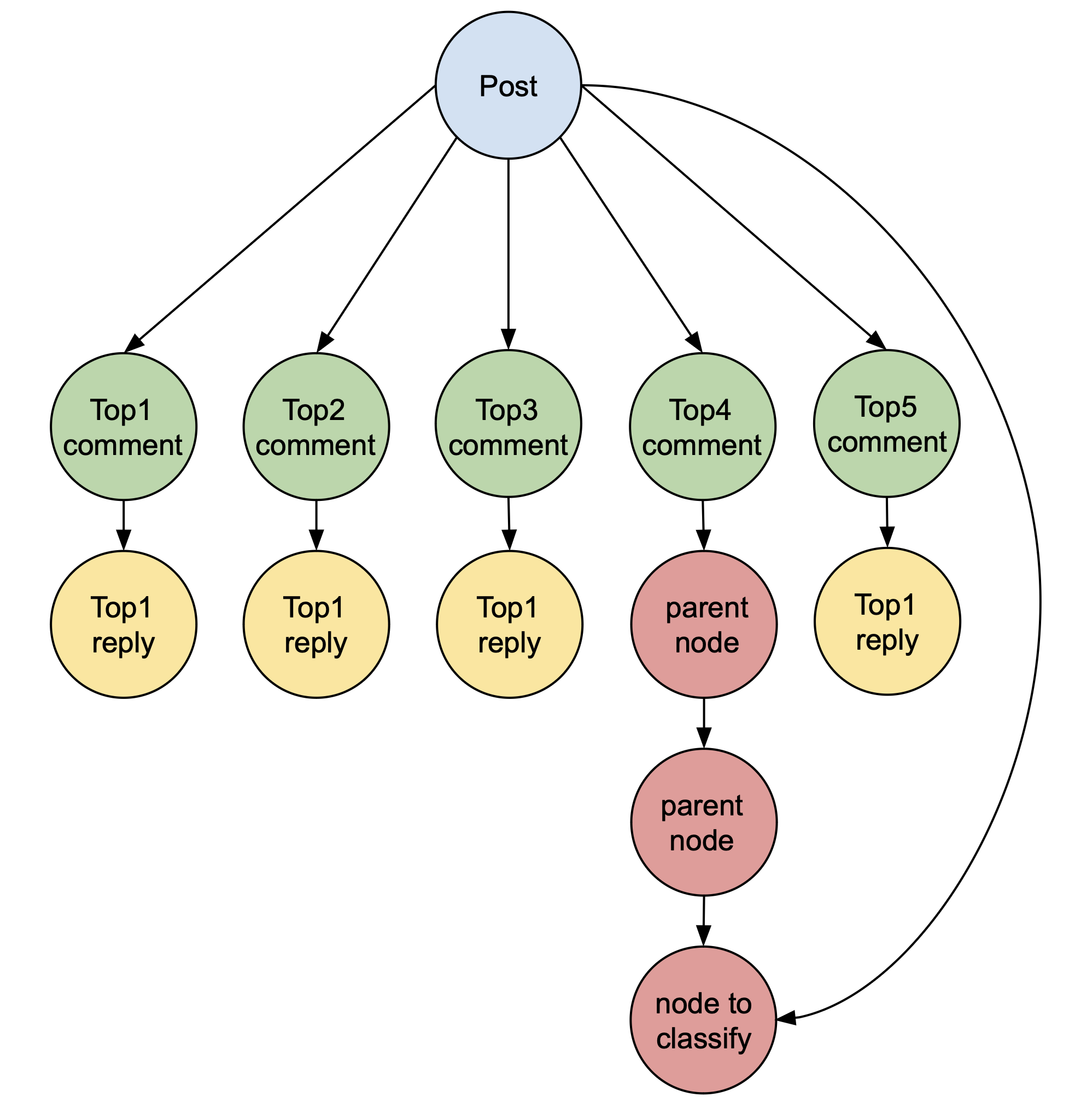} 
  \caption {Diagram of a Reddit conversation processed using the affordance-based method. The left figure illustrates the edge option used in the paper (\textit{\(trim_{final}\)}), while the right one represents the alternative option with fewer edges (\textit{\(trim_{alt}\)}).}
\label{fig:trim-strat}
\end{figure*}

For affordance-based graphs, we explored two edge configurations to account for the assumption that users read the initial post before commenting. The first approach, \(trim_{final}\), connects the post node to all other nodes, while the alternative, \(trim_{alt}\), links the post only to the target node. Figure~\ref{fig:trim-strat} illustrates these configurations.  

Table~\ref{tab:trim-strat} reports the F1-scores for different trimming strategies, evaluated using the 3-layer GAT architecture (Section~\ref{subsec:model-archi}). The affordance-based method with \(trim_{final}\) achieved the highest performance and was selected for all experiments.  

\begin{table}[h!]
    \centering
    \small
    \begin{tabular}{|c|c|}
        \hline
        \textbf{Trimming Strategy} & \textbf{F1 Score} \\
        \hline
        Affordance with \(trim_{final}\) & 0.765 \\
        Affordance with \(trim_{alt}\) & 0.761 \\
        Recent & 0.758 \\
        \hline
    \end{tabular}
    \caption{F1-scores for different trimming strategies.}
    \label{tab:trim-strat}
\end{table}

\subsection{Objective Function}
\label{subsec:objective-function}

The classification task is formulated as estimating the conditional probability distribution:  
\[
P\left(y_i \mid G_i, \{u_j : c_j \in V_i\}\right),
\]  
where \( \{u_i : c_i \in V_t\} \) denotes the textual content of comments within the subgraph \( G_t \). The model parameters \( \theta \) are learned by minimizing the binary cross-entropy loss:

\begin{equation}
\resizebox{\columnwidth}{!}{$
\mathcal{L}(\theta) = - \frac{1}{|D|} \sum_{(G_t, y_t) \in D} \Big[ y_t \log f_\theta(G_t, \{u_i\}) \\
+ (1 - y_t) \log \left(1 - f_\theta(G_t, \{u_i\})\right) \Big],
$}
\end{equation}

where \( D \) represents the training dataset, and \( f_\theta \) is the graph-based classification model that integrates both structural and textual features.

\subsection{Experiment Setup}
\label{subsec:experiment-setup}

\paragraph{Dataset Adaptation}

We evaluate models on the augmented and balanced Contextual Abuse Dataset (CAD) (Section~\ref{sec:dataset}). To reconstruct conversation trees for labeled comments, we use Arctic-Shift \citep{arcticshift2024}, an API tool for retrieving past Reddit data. Due to class imbalance (81.2\% non-abusive vs. 17.8\% abusive), we apply under-sampling, resulting in a balanced dataset of 7,210 samples (3,605 per class). All abusive labels are merged into a single "abusive" category, and all non-abusive labels into "non-abusive." The dataset is split into 80\% training (5,768 samples), 10\% validation (721 samples), and 10\% test (721 samples), ensuring class balance.

\paragraph{Hyperparameter Tuning and Model Training}

The model was trained with a learning rate of $3 \times 10^{-6}$, weight decay of 0.1, a dropout rate of 0.3 for BERT embeddings, and 0.4 for GAT layers. Due to memory constraints, training used a true batch size of 1 with gradient accumulation over 16 steps, resulting in an effective batch size of 16. Early stopping was applied after seven epochs without improvement. 

The graph-based model incorporates a BERT-base encoder (110M parameters) alongside 1 to 5 layers of Graph Attention Networks (GAT). Each GAT layer introduces approximately six million additional parameters due to multi-head attention, resulting in total model sizes ranging from 116M (1-layer GAT) to 140M (5-layer GAT). All models were trained on trimmed conversation graphs to ensure consistency, with ten seeded runs for reproducibility. We report mean F1 scores with 95\% confidence intervals, assuming a two-tailed normal distribution (\(n=10\), \(t=2.228\)).

Training was performed on Nvidia RTX 8000 GPUs and on Nvidia H100 GPUs (96 GB each). The training time for a full 20-epoch run ranged from approximately 10 to 16 hours depending on the model, and hardware configuration.

\paragraph{Implementation Details} We implemented our method using PyTorch \citep{paszke2019pytorch} and PyTorch Geometric \citep{fey2019fast}. We initialized with pre-trained BERT weights, and utilized the Hugging Face Transformers library \citep{wolf2020transformers}. For optimization, we employed the AdamW optimizer \citep{loshchilov2019decoupled}.  

\end{document}